\documentclass[runningheads]{llncs}

\usepackage{eccv}

\usepackage{eccvabbrv}
\usepackage[sectionbib]{chapterbib}
\usepackage{amsmath}
\usepackage{graphicx}
\usepackage{textcomp}
\usepackage{xcolor}
\usepackage{makecell}
\usepackage{multirow}
\usepackage{float}
\usepackage{graphicx}
\usepackage{booktabs}

\usepackage[accsupp]{axessibility}  
\definecolor{pinkishmagenta}{RGB}{221, 35, 142} 

\usepackage{hyperref}
\usepackage{wrapfig}
\usepackage{algorithm2e}
\RestyleAlgo{ruled}
\usepackage{amssymb}
\usepackage{pifont}
\usepackage{orcidlink}
\newcommand{\cmark}{\ding{51}}%
\newcommand{\xmark}{\ding{55}}%

\begin{document}

\title{CLAMP-ViT: Contrastive Data-Free Learning for Adaptive Post-Training Quantization of ViTs} 

\titlerunning{CLAMP-ViT}
\author{Akshat Ramachandran\inst{1} \orcidlink{0009-0000-4763-3321} \and
Souvik Kundu\inst{2} \orcidlink{0000-0002-3533-9405} \and
Tushar Krishna\inst{1} \orcidlink{0000-0001-5738-6942}}

\authorrunning{Akshat Ramachandran, Souvik Kundu, and Tushar Krishna}

\institute{Georgia Institute of Technology, Atlanta, GA, USA \\
\email{akshat.r@gatech.edu, tushar@ece.gatech.edu} \and
Intel Labs, San Diego, CA, USA \\
\email{souvikk.kundu@intel.com}}

\maketitle
\setcounter{footnote}{0}
\begin{abstract}
We present CLAMP-ViT, a  data-free post-training quantization method for vision transformers (ViTs). We identify the limitations of recent techniques, notably their inability to leverage meaningful inter-patch relationships, leading to the generation of simplistic and semantically vague data, impacting quantization accuracy. CLAMP-ViT employs a two-stage approach, cyclically adapting between data generation and model quantization. Specifically, we incorporate a patch-level contrastive learning scheme to generate richer, semantically meaningful data. Furthermore, we leverage contrastive learning in layer-wise evolutionary search for fixed- and mixed-precision quantization to identify optimal quantization parameters while mitigating the effects of a non-smooth loss landscape. Extensive evaluations across various vision tasks demonstrate the superiority of CLAMP-ViT, with performance improvements of up to 3\% in top-1 accuracy for classification, 0.6 mAP for object detection, and 1.5 mIoU for segmentation at similar or better compression ratio over existing alternatives. Code is available at \url{https://github.com/georgia-tech-synergy-lab/CLAMP-ViT.git}

\end{abstract}
\section{Introduction}
\label{sec:introduction}
Vision transformers \cite{dosovitskiy2020image} (ViTs) have recently gained a lot of traction due to their state-of-the-art (SoTA) performance across various computer vision (CV) tasks \cite{ramachandran2023ntrans, zhang2021vit, strudel2021segmenter, zhang2023completionformer, ansari2023gpu}. Concurrently, the growing need to deploy these parameter-heavy models at the resource-limited edge \cite{frumkin2023jumping}, has inspired research on various model compression techniques.
\begin{wrapfigure}{R}{0.50\textwidth}
\vspace{-12mm}
  \centering \includegraphics[width = 0.50\textwidth]{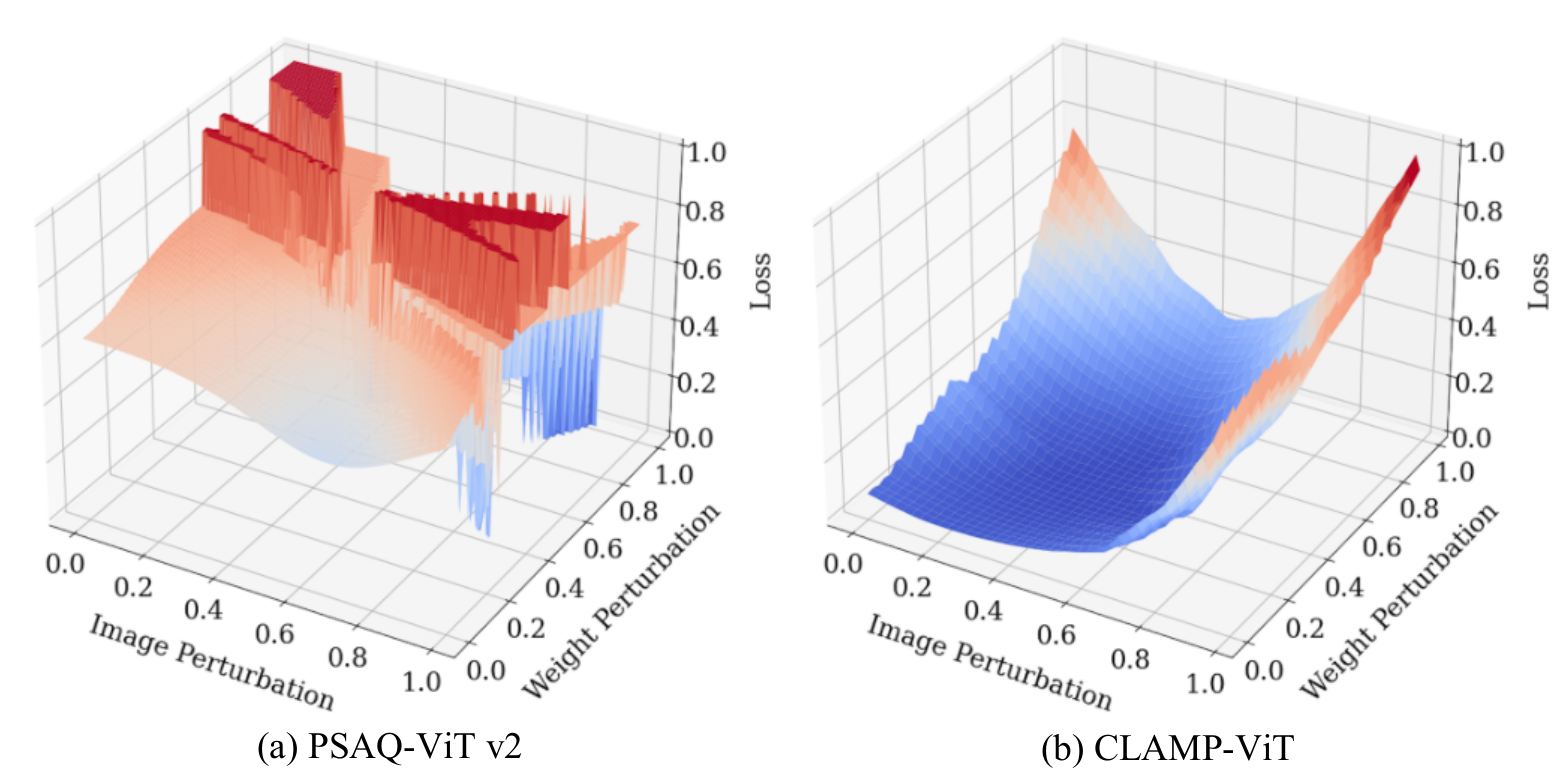}
  \caption{Visualization of loss landscape of (a) PSAQ-ViT v2 and (b) CLAMP-ViT on DeiT-S with perturbations to quantized model weights and synthetic data \cite{li2018visualizing}.}
  \vspace{-8mm}
  \label{fig:teaser}
\end{wrapfigure}
Model \emph{quantization} \cite{ranjan2024lrp, lin2021fq, frumkin2023jumping, yuan2022ptq4vit} has emerged as a popular technique to achieve memory and compute efficient deployment. Quantization reduces memory footprint and improves computation speed of a model by mapping full-precision (FP) weights to reduced precision formats (e.g., $\leq 8$-bit INT) \cite{zhang2018lq, hubara2016binarized, kim2020exploiting, fei2021general}. In particular, quantization-aware training (QAT) allows a model to train by taking quantization approximation in to account, enabling ease of quantization. Post-training quantization (PTQ), in contrast, acts as a \textit{plug-and-play} quantization applied on a pre-trained model. PTQ has become popular as it can leverage the pre-trained model and does not add training overhead \cite{li2023vit, li2022q, kundu2022bmpq}. We thus consider the PTQ setting in this work. However, PTQ requires access to a calibration set that is often drawn from the training data \cite{hubara2021accurate, yuan2022ptq4vit}. This may be infeasible in situations involving privacy and security concerns \cite{kundu2021analyzing, zhang2023sal}, making these techniques ill-suited to yield optimal performance.

Recent works \cite{li2023psaq, li2022patch} propose data-free PTQ (DFQ), generating synthetic calibration data $T_{\text{syn}}$ from Gaussian noise \cite{choi2021qimera, li2022patch}, embedding information from the original dataset $T_{\text{orig}}$, where $T_{\text{syn}} << T_{\text{orig}}$.

Existing DFQ methods for CNNs \cite{cai2020zeroq, zhong2022intraq} exploit the batch-normalization (BN) layer statistics \cite{peters2018probabilistic, yin2020dreaming} to generate synthetic samples mimicking the original data distribution. For ViTs, absence of the BN layer makes these techniques obsolete. Recent efforts to extend DFQ to ViTs include PSAQ-ViT v1 \cite{li2022patch} and PSAQ-ViT v2 \cite{li2023psaq}. They rely on information embedded in the attention score output of the multi-head self attention (MHSA) layer. PSAQ-ViT v1 and v2 \cite{li2022patch, li2023psaq} introduce a relative value metric namely, \textit{patch similarity}, to optimize Gaussian noise towards synthetic data by maximizing the entropy of patch similarity in a global manner.
%
%
However, the metric considered in PSAQ-ViT v1 and v2 assumes all patches\footnote{Patch (subset of image): group of neighboring pixels in an image.}to be \textit{equally} important, without considering spatial sensitivity \cite{zhang2023patch}. This may fail to capture semantically meaningful inter-patch relations, potentially affecting robustness of the synthetic data. As we can see in \cref{fig:teaser}(a) even insignificant perturbations in the generated images (augmenting pixels) or weights (to simulate quantization process) may cause significant jaggedness in the loss landscape of PSAQ-ViT v2. This also implies that the predictions may have large discrepancy even for small input/weight perturbation. Moreover, the synthetic data generation stage in these methods does not consider the informativeness of the generated samples towards the quantization process, nor do they establish countermeasures to ameliorate the non-smooth loss landscape during quantization, resulting in sub-optimal parameter search and poor generalization to test set \cite{chen2022bootstrap}.

\noindent
\textbf{Our contributions. }The discussion above hints at the potential limitation of \cite{li2023psaq, li2022patch} in capturing semantically meaningful and robust inter-patch relationships to generate synthetic data that is well-suited to quantization. Towards solving these limitations, we present \textit{contrastive data-free learning for adaptive post-training quantization of ViTs} (\textbf{CLAMP-ViT}), a general DFQ method applicable to a wide range of vision tasks. To the best of our knowledge, CLAMP-ViT is the \textbf{first work to support both fixed- \footnote{weights/activations quantized to same precision for all layers.} and mixed-precision \footnote{weights/activations quantized to different precision for different layers.} DFQ of ViTs}, starting from a pre-trained FP model. 

Specifically, CLAMP-ViT utilizes the architectural characteristics of ViTs and inherent properties of real-images to generate semantically rich and meaningful data while ensuring spatial sensitivity. 
Here we leverage a novel patch-level contrastive learning scheme, where for each patch (anchor) in the MHSA layer output, we treat semantically similar patches in a neighborhood around the patch as positive patches (evaluated using cosine similarity) and the remaining patches in the neighborhood as negative patches (see \cref{fig:method_patch} for an intuitive visualization). We then leverage a patch-level contrastive loss that drives the representation of the anchor patch closer to the positive patches and away from the negative patches, enabling exploration of semantically meaningful relations. 

Recently, Evol-Q \cite{frumkin2023jumping} identified a limitation of gradient based methods \cite{li2023psaq} to search over quantization parameters of ViTs having non-smooth loss landscape, leading to poor accuracy and generalization. We take inspiration from this and a recent work \cite{ramachandran2024algorithm} to design a layer-wise evolutionary search to identify the optimal bit-width and scale factors for each layer. Additionally, we propose a novel local contrastive objective to capture distributional variance in intermediate layer outputs (crucial for layer-wise quantization) and improve search convergence. This loss sufficiently captures the representational divergence of intermediate layers outputs to identify optimal quantization parameters and yields a smooth loss landscape, as demonstrated in Fig. \ref{fig:teaser}(b), enabling generalizability and better performance on test data \cite{chen2022bootstrap}. Furthermore, to ensure the generated data is adaptive to the quantization process, CLAMP-ViT performs a \textit{cyclically adaptive} strategy alternating between data-generation and quantization.

To evaluate the merits of CLAMP-ViT we conduct extensive experiments on image classification, detection, and segmentation with different ViT variants and quantization scenarios and observe superior performance over SoTA.

\section{Related Works}
\label{sec:related_work}

\textbf{Data-Driven PTQ for ViTs. }PTQ offers an efficient alternative to QAT \cite{li2022q, li2023vit} by directly quantizing pre-trained models without the need for compute-heavy retraining. In specific, PTQ4ViT \cite{yuan2022ptq4vit} employs twin uniform quantization and Hessian-guided scale optimization. FQ-ViT \cite{lin2021fq} uses a power-of-two factor quantization to handle inter-channel variation in LayerNorm and log-INT-Softmax. RepQ-ViT \cite{li2023repq} separates quantization and inference, optimizing accuracy and efficiency by employing scale-reparameterized quantizers. These methods apply fixed-precision quantization, assuming all layers support similar approximations, potentially leading to sub-optimal accuracy \cite{ranjan2024lrp}. On the other hand, techniques like, VT-PTQ \cite{liu2021post} adopt mixed precision for specific modules based on sensitivity metrics, while PMQ \cite{xiao2023patch} and LRP-QViT \cite{ranjan2024lrp} allocate bit-widths by assessing both layer sensitivity and contribution to the output, respectively. However, all these methods assume a part of training set to be available for calibration which may be infeasible due to privacy \cite{kundu2021analyzing}.

\noindent
\textbf{Data-Free PTQ for ViTs. }
Recent data-free PTQ efforts (DFQ) for ViTs including PSAQ-ViT v1 \cite{li2022patch} and v2 \cite{li2023psaq} utilize a "patch similarity" metric to refine Gaussian noise to synthetic data resembling real images. In specific,  both these approaches leverage the fact that the self-attention has different response to real image and noise. PSAQ-ViT v1 \cite{li2022patch} uses a two-stage cascaded framework to generate synthetic data for model quantization on image classification tasks. PSAQ-ViT v2 \cite{li2023psaq} expanded this to a broader range of tasks, employing a MinMax game between full precision and quantized models for data generation and quantization. Despite these innovations, both the versions face several limitations as highlighted in \cref{sec:introduction}, leading to poor DFQ performance, particularly at reduced precision. Additionally, these methods support only fixed-precision quantization. CLAMP-ViT, in contrast, introduces a DFQ method that addresses their shortcomings in yielding SoTA performance, while supporting both fixed and mixed-precision quantization.


\noindent
\textbf{Contrastive Learning. } Contrastive learning as a technique has been widely adopted in self-supervised settings \cite{fu2022contrastive, cao2022synergistic} and is proven to combat overfitting via regularization against negative samples. Recently, Evol-Q \cite{frumkin2023jumping} demonstrated the benefits of a global contrastive objective \cite{chuang2022robust, yeh2022decoupled} coupled with evolutionary search in smoothening the loss landscape while calibrating the scaling factors of a pre-quantized model to enhance accuracy. Unlike \cite{frumkin2023jumping} that only aims to adjust the scale factors of each layer of a pre-quantized model, we conduct complete quantization starting from a FP model. On evaluation of the global contrastive objective used in Evol-Q, we find it to be sub-optimal, causing premature convergence while quantizing an FP model. Instead, we present a local contrastive loss that captures the distributional variance in intermediate layer outputs, drastically improving convergence and hence, identification of quantization parameters.

\section{CLAMP-ViT Framework}
\begin{figure}[t]
  \centering \includegraphics[width = 0.95\columnwidth, keepaspectratio]{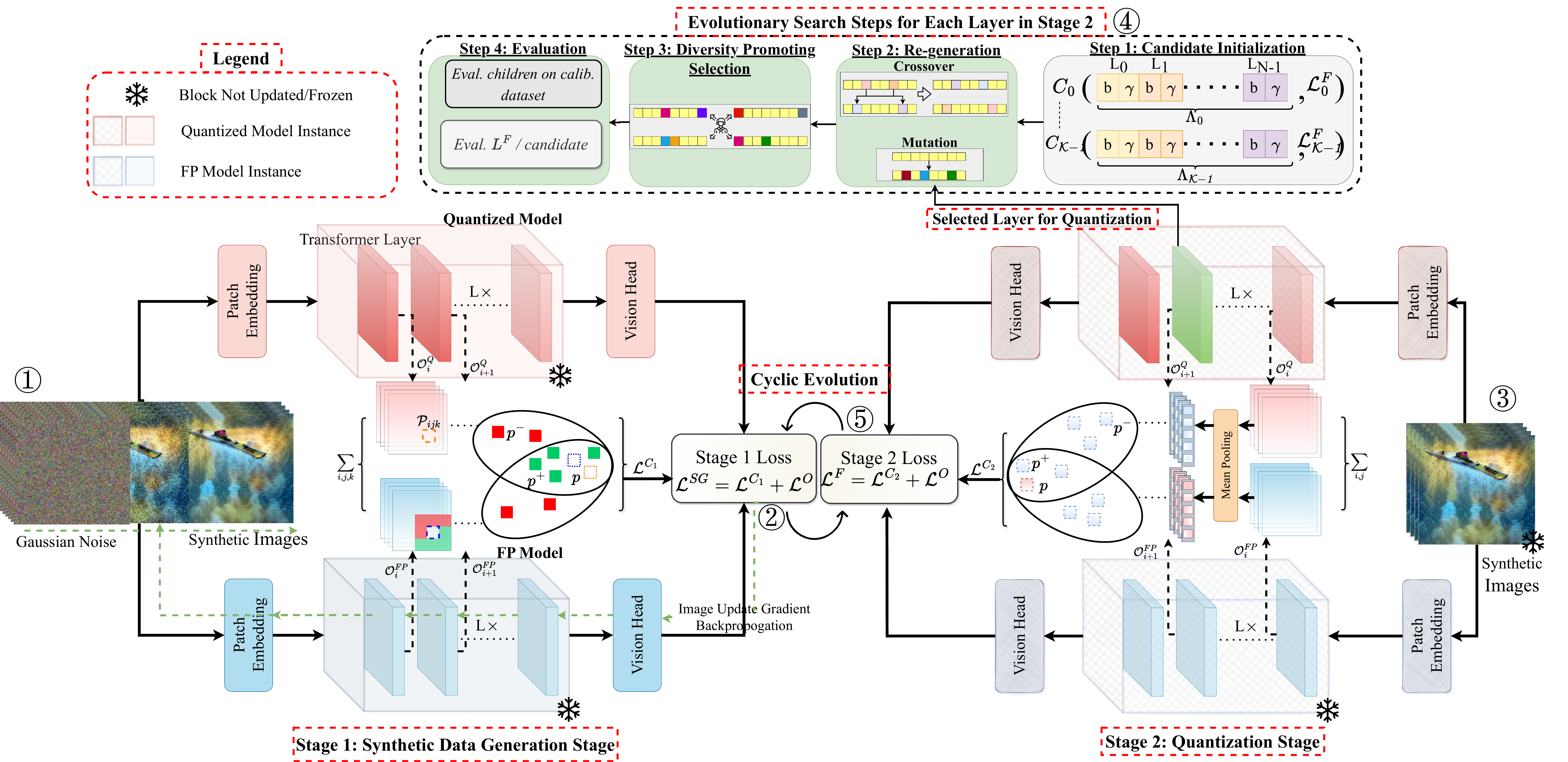}
  \caption{\textbf{Overview of the cyclically evolving two-stage CLAMP-ViT framework}. In stage 1 (\textcircled{1} - \textcircled{2}), $\mathcal{L}^{SG}$ is minimized to update Gaussian noise towards synthesizing data. Stage 2 (\textcircled{3} - \textcircled{5}) conducts layer-wise evolutionary search to identify optimal quantization parameters while minimizes $\mathcal{L}^F$. Illustrated with multiple instances of models for clarity, only one instance of each model is used in the framework.}
  \label{fig:method_clamp}
  \vspace{-5mm}
\end{figure}

We present an overview of CLAMP-ViT in \cref{fig:method_clamp}. In this section, we first go through notations, computational process of ViTs and quantization strategy in the preliminaries, followed by a detailed description of the proposed contrastive loss and the two stages of CLAMP-ViT. Finally, the overall DFQ pipeline is summarized and presented (refer to  Supplementary for the detailed Algorithm).

\subsection{Preliminaries}
\label{sec:prelim}

\textbf{Notations. }Let ${X} \in \mathcal{R}^{H \times W \times C}$ be the input image to an $L$-layer ViT, where $(H, W, C)$ are the height, width, and channels, respectively (we ignore the batch dimension for simplicity). The input is partitioned into  ${N}$ non-overlapping patches that are then linearly transformed to $d$-dimensional patch embeddings, ($\mathcal{R}^{N \times d}$) that is passed through an encoder consisting of a series of transformer layers each composed of an MHSA and an MLP module. Each MHSA module is composed of $h$ heads, to capture long-range patch correlations \cite{li2022efficientformer}. At head $j$, ${X}$ is transformed to query ($Q_j$), key ($K_j$), and value ($V_j$) tensors to perform self-attention $\Phi_j$, that is linearly projected after concatenation over the heads,
\vspace{-2mm}
\begin{equation}
    \Phi_j(Q_j, K_j, V_j) = \textbf{softmax}(Q_jK_j^T/\sqrt{d})V_j ,
\end{equation}
\vspace{-4mm}
\begin{equation}
    \text{MHSA}(Q, K, V) = \textbf{concat}(\Phi_0, \Phi_1, ... \Phi_{h-1})W^0
\end{equation}
\noindent
The output of the $j^{th}$ head at $i^{th}$ MHSA layer is given by $O_{i,j} \in \mathcal{R}^{N \times d}$. The series of transformer layers are succeeded by task-specific heads for classification, detection, or segmentation, depending on the nature of the vision task.

\noindent
\textbf{Quantization. }In this paper, we perform uniform symmetric quantization (fixed-/mixed-precision) of both weights and activations for ViTs, mapping full precision values into a uniform scale determined by bit-width (b), given  as,
\begin{equation}
    Q(X, \gamma, b) = \emph{clip}(\Bigl\lfloor{\frac{X}{\gamma}}\Bigr\rceil, -2^{b-1}+1, 2^{b-1} -1)
\end{equation}
\noindent
where $\gamma$ is the scale factor and $X$ represents the FP tensor.

\subsection{Contrastive Objective}
\label{sec:contrastive_prelim}

Contrastive learning based on the infoNCE loss \cite{wang2023positive} helps learn an anchor sample from both the similar (positive) and dissimilar (negative) samples, typically  using a softmax function to normalize the similarities into probabilities. However, infoNCE loss suffers from a disproportionate impact of  a single positive and many negative samples \cite{frumkin2023jumping}. This can affect learning the synthetic data as well as the quantized model parameters. Inspired by \cite{wang2023positive}, we present a modified infoNCE loss ($\mathcal{L}^C_{i,j}$ in Eq. \ref{equation:contrastive}) that addresses this imbalance yielding latent impact of positive and negative samples pairs equally. 
\begin{equation}
\label{equation:contrastive}
    \mathcal{L}^C_{i,j} = -\log \frac{ \sum_{p+}\exp(\lambda^p_{i,j} \cdot \lambda_{i,j}^{p+} / \tau)}{ \sum_{p+} \exp (\lambda^p_{i,j} \cdot \lambda_{i,j}^{p+} / \tau) + \sum_{p-} \exp(\lambda^p_{i,j} \cdot \lambda_{i,j}^{p-} / \tau)}
\end{equation}

\noindent
here superscript $p$, $p+$, and $p-$ correspond to the anchor prediction (drawn from quantized model), positive, and negative prediction (drawn from FP model), respectively. $\tau$ controls the concentration level \cite{wang2023positive}. Subscript $(i, j)$  represents (\# layer id, \# head id), and (\#layer id, \#batch id) for the data generation and quantization stage, respectively. $\lambda$ represents the variable taken to evaluate the log likelihood-- the intermediate layer patch embeddings (stage 1) or layer output activations (stage 2). The final stage $s$ loss is given as $\mathcal{L}^{C_s} = \sum_i\sum_j\mathcal{L}^{C_s}_{i,j}$.

\begin{wrapfigure}{R}{0.53\textwidth}
  \vspace{-7mm}
  \centering \includegraphics[width = 0.53\textwidth]{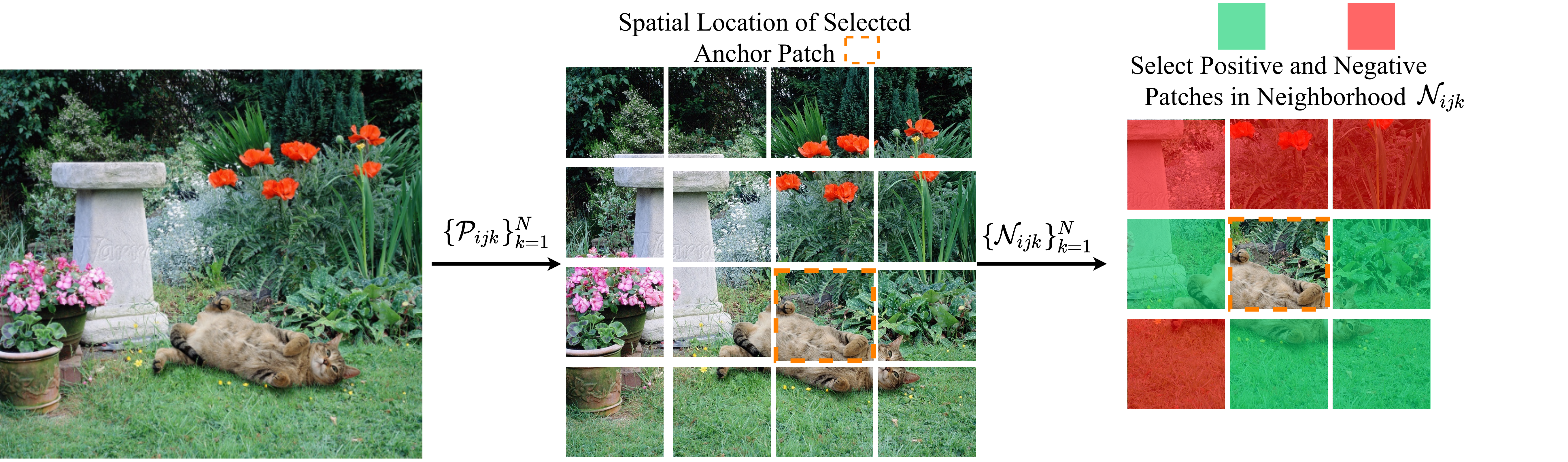}
  \caption{Intuitive visualization of positive and negative patch selection in Stage 1.}
  \vspace{-8mm}
  \label{fig:method_patch}
\end{wrapfigure}
\label{sec:clamp_vit}

\subsection{Stage 1: Synthetic Data Generation}
Our goal in Stage 1 is to generate semantically rich and meaningful images that can reliably exploit inter-patch relations while ensuring informativeness to the quantization process. For each "\textit{anchor patch}" of an MHSA output in the quantized model, we first locate positive and negative patches (see \cref{fig:method_patch}) from same layer id in the FP model, from a neighborhood of the anchor's spatial position. We then leverage a contrastive learning objective (see \cref{sec:contrastive_prelim}) that operates to maximize the similarity between the anchor and positive patches while minimizing similarity with the negative patches. Selection of the anchor patch from the quantized model for guiding data generations helps the model to recognize the semantic context within the generated data, during the subsequent quantization stage (\textit{informativeness}).

\noindent
\textbf{Patch Neighborhood. } For every $k^{th}$ anchor patch $\mathcal{P}_{ijk}$ corresponding to the ${j}^{\text{th}}$ head in ${i}^{\text{th}}$ MHSA layer of the quantized model, we identify a neighborhood of size $\mathcal{N}_{ijk}$ with  the same spatial location of $\mathcal{P}_{ijk}$ as the center, but located in the MHSA layer output of the FP model. Within $\mathcal{N}_{ijk}$, to identify the most semantically correlated patches ($\mathcal{P}_\mathcal{N} \in \mathcal{N}_{ijk}$) we use cosine similarity as follows,    

\begin{equation}
    \rho(i,j,k) = \frac{\mathcal{P}_{ijk}^T \cdot \mathcal{P}_{\mathcal{N}}}{||\mathcal{P}_{ijk}||_2 \cdot || \mathcal{P}_{\mathcal{N}} ||_2}
\end{equation}

\noindent
The cosine similarity $\rho$ is estimated $\forall P_\mathcal{N} \in \mathcal{N}_{ijk}$. We then select the positive patches to be the top-$n$ patches that have the highest $\rho(i,j,k)$ with the anchor patch and the rest of the patches in the neighborhood correspond to negative patches. We empirically set $n = 4$ in our experiments for a neighborhood of size $3 \times 3$. We then compute $\mathcal{L}^{C_1}_{i,j}$ for all anchor patches for each attention head output over all layers to get the contrastive loss ($\mathcal{L}^{C_1}$). We then compute the sample generation loss $\mathcal{L}^{SG}$ by combining $\mathcal{L}^{C_1}$ and the \textit{mean absolute error} (MAE) (see \cref{sec:pipeline}) output loss ($\mathcal{L}^O$) and minimize it for data generation.

\subsection{Stage 2: Quantization}
\label{sec:evolutionary}
We now present our layer-wise evolutionary search with a local contrastive loss-based fitness function to rank suitable quantization parameters from a large search space. We detail the fitness function and the search steps as follows.

\noindent
\textbf{Fitness Function. }To evaluate the quantization parameters explored by the evolutionary search algorithm, we introduce a fitness function $\mathcal{L}^F$ that combines the contrastive loss $\mathcal{L}^{C_2}$ (\cref{equation:contrastive}) and the MAE loss ($\mathcal{L}^O$) computed with respect to the targets at the output (explained in \cref{sec:pipeline}) i.e, $\mathcal{L}^F = \mathcal{L}^{C_2} +\mathcal{L}^O$. Unlike Stage 1, we use the intermediate activations after each transformer layer to compute the contrastive loss $\mathcal{L}^{C_2}$. For a ViT, let the set of intermediate representations of FP and quantized model be denoted as, $\mathcal{O}^{fp} = \{\mathcal{O}^{fp}_0, \mathcal{O}^{fp}_1, .... \mathcal{O}^{fp}_{L-1}\}$  and $\mathcal{O}^{q} = \{\mathcal{O}^{q}_0, \mathcal{O}^{q}_1, .... \mathcal{O}^{q}_{L-1}\}$, respectively. However, directly using high dimensional $\mathcal{O}^{fp}$ and $\mathcal{O}^{q}$ can result in high compute overhead. Therefore for each layer $i$, at the intermediate output, we perform mean pooling along the patch dimension to obtain a low-dimensional representation ($\mathcal{O}_{i} \in \mathcal{R}^N$), reducing it by a factor of $h\times d$. We then apply the contrastive loss (\cref{equation:contrastive}) sampled within the batch dimension on the low-dimensional $\mathcal{O}_{i}^{fp}$ and $\mathcal{O}_{i}^{q}$. For the layer output activation generated from each constituent of a batch of synthetic data, the anchor ($\lambda^p$) corresponds to the intermediate layer output of the quantized model, positive ($\lambda^{p+}$) and negative ($\lambda^{p-}$) samples correspond to set of intermediate layer outputs of FP model that have the same and different targets respectively, relative to the anchor within the batch.     

\noindent
\textbf{Step 1: Candidate Initialization. }A candidate quantization solution is encoded as a set $\Lambda$ of $L$ tuples, such that for layer $i$, tuple $\Lambda[i]$ represents the two quantization parameters $\langle b, \gamma \rangle$.
 $b$ can take any integer value between $2$ and $8$, while $\gamma$ is constrained to a uniform ball of radius $10^{-3}$ centered around, $\gamma[i]^{init}$\footnote{$(\textbf{max}(\theta_i)-\textbf{min}(\theta_i))/(2^b - 1) $, where $\theta$ is the weight tensor.}. The candidate scale factors are sampled as $\gamma[i] = \gamma[i]^{init} + f(-10^{-3}, +10^{+3})$, where $f$ is a random sampling function. We begin the evolutionary search by creating a population via randomly sampling $\mathcal{K}$ candidate $\Lambda$s, each consisting of layer-wise quantization parameters. We then evaluate the fitness function $\mathcal{L}^F$ for each candidate $\Lambda$. We create $\mathcal{K}$ tuples each with $(\Lambda, \mathcal{L}^F)$ to form the initial population. $\mathcal{L}^F$ of each initial candidate with corresponding set $\Lambda$ is pre-computed and stored to avoid recomputation.

\noindent
\textbf{Step 2: Re-generation (Crossover and Mutation). } Each candidate in the population is ranked based on  corresponding $\mathcal{L}^F$s (lower the better) of which the top two serve as the parents for the next candidate generation (child). When evolving candidates, perturbing too many layer parameters based on parents can lead to search instability. To mitigate this, at each evolution step we employ a layer-wise regeneration approach, evolving a single transformer layer at a time based on chosen parents, keeping all other layer parameters to that of the top-1 parent (ranked via $\mathcal{L}^F$). The child's parameter regeneration for a layer based on that of the chosen parents ($p1$, $p2$) is formulated as,
    \begin{gather}
    \label{eq:regeneration}
        b_{{child}} = \mathbf{random}(\mathbf{min}(b_{p1}, b_{p2})-1, \mathbf{max}(b_{p1}, b_{p2}) + 1) \\
        \label{eq:regeneration_2}
        \gamma_{{child}} = \mathbf{mean}(\gamma_{p1}, \gamma_{p2}) + \eta(-10^{-3}, 10^3)
    \end{gather}

\noindent
\textbf{Step 3: Diversity Promoting Selection. } To avoid overfitting during search we follow \cite{dong2023emq} and introduce diversity into the population. In specific, we create `$P=5$' random parents and use each of them to crossover with the child generated in Step 2 and create `$P$' diverse children by following \cref{eq:regeneration}, \cref{eq:regeneration_2}.

\noindent
\textbf{Step 4: Evaluation and Population Update. } We evaluate all generated children in the steps above and acquire $\mathcal{L}^F$s. The child generated in Step 2 and the corresponding fitness function value is added to the population. We then rank the diversity promoting children from Step 3 and select the best child to be added to the population for the next iteration.
In our layer-wise evolutionary search strategy, we employ $\mathcal{P}$ passes over all layers of a ViT, and each layer is iterated over $\mathcal{C}$ cycles in each pass. So, the population is updated $\mathcal{P} \times \mathcal{C} \times L$ times, i.e., the Step 2, 3, and 4 are iteratively executed $\mathcal{P} \times \mathcal{C} \times L$ times.

\noindent
\textbf{Activation Quantization. }We note that the sensitivity to quantization for activations is closely correlated to the sensitivity of weight parameters. Therefore for layer $i$, we determine the output activation quantization parameters as follows, $b_{act}[i]$ = $\textbf{min}(8, b[i] \times 2)$ and $\gamma_{act}[i] = \gamma_{act}[i-1] + \gamma[i]$.


\subsection{Overall Pipeline}
\label{sec:pipeline} 
In \cref{fig:method_clamp}, we illustrate the complete CLAMP-ViT framework. To ensure adaptive data-generation and informativeness for quantization, we use a cyclic strategy between the two stages, updating generated data based on the quantized model's needs for optimal parameter search. The framework requires an input batch of $\mathcal{B}$ random Gaussian images $X_{\mathcal{B}}$, and corresponding task-specific targets $T_{G_{\mathcal{B}}}$ ($T_{G_{\mathcal{B}}}$ for each task is detailed in \cref{sec:results}). The targets direct the synthetic image generation towards task-specific goals as well as penalize inaccurate predictions of quantized model through the output loss $\mathcal{L}^O$,

\begin{equation}
    \mathcal{L}^O = \frac{1}{n_c}(||\mathcal{Q}(X_{\mathcal{B}}) - T_{G_{\mathcal{B}}}||_1 + ||\mathcal{FP}(X_{\mathcal{B}}) - T_{G_{\mathcal{B}}}||_1) 
\end{equation}

\noindent
where $n_c$ is the number of output classes (classification) or prediction map size (segmentation/detection). The quantized model is initialized to the best candidate from $\mathcal{K}$ tuples. The framework assumes a range of bit-widths and a single bit-width for mixed- and fixed-precision search, respectively. In Stage 1, $X_{\mathcal{B}}$ is fed to the frozen quantized ($\mathcal{Q}$) and full-precision model ($\mathcal{FP}$) to minimize the sample generation loss $\mathcal{L}^{SG} = \mathcal{L}^{C_1} + \mathcal{L}^O$, updating $X_{\mathcal{B}}$ via backpropagation for $G$ iterations. In Stage 2, we use the generated data to  quantize  $\mathcal{Q}$ for $\mathcal{P} \times \mathcal{C} \times L$ iterations by minimizing $\mathcal{L}^F$. We cyclically update the generated data every $\mathcal{C}/2$ iterations. In every subsequent execution of Stage 1, we do not restart from Gaussian noise but use $X_{\mathcal{B}}$ from the previous Stage 1 execution and update it for $G/2$ iterations. In this manner, the two stages are executed alternately. 

\section{Experimental Results}
\label{sec:results}
\subsection{Experimental Setup}
\label{sec:setup}


\begingroup  
\begin{table}[t]\centering
 \caption{Fixed-precision quantization accuracy comparison with SoTA on image classification tasks with ImageNet-1k testset. `R', `S' signifies real and synthetic calibration data and W/A indicates weight/activation bit-width. The values in \textbf{bold} and \underline{underline} signifies, the best performance overall, and with synthetic data, respectively.}
 \vspace{-10pt}
  \renewcommand*{\arraystretch}{1.0}
  \setlength\tabcolsep{1.9pt}
\resizebox{0.7\linewidth}{!}
{%
\begin{tabular}{cccc|cc|cc}
\Xhline{2\arrayrulewidth}
Model & Method & Data & \#Images &  W/A & Top-1 & W/A & Top-1 \\
\Xhline{2\arrayrulewidth}
  \multirow{6}{*}{ViT-B}& Baseline & - & - & 32/32 & 84.53 & 32/32 & 84.53 \\ \cline{2-8}
 & PSAQ-ViT v1 \cite{li2022patch} & S & 32 & 8/8 & 37.36 & 4/8 & 25.34 \\
  & PTQ4ViT \cite{yuan2022ptq4vit} & R & 32 & 8/8 & \textbf{84.25} & 4/8 & 67.16 \\
  & FQ-ViT \cite{lin2021fq} & R & 1000 & 8/8 & 83.31 & 4/8  & \textbf{78.73} \\
 & RepQ-ViT \cite{li2023repq} & R & 32 &  8/8 & 81.45 & 4/8  & 76.29 \\
 & \textbf{CLAMP-ViT (Ours)} & S & 32 & 8/8 & \underline{84.19} & 4/8 & \textbf{\underline{78.73}} \\
 \Xhline{1\arrayrulewidth}
 
 \multirow{6}{*}{DeiT-T}& Baseline & - & - & 32/32 & 72.21 & 32/32 & 72.21 \\ \cline{2-8}
 & PSAQ-ViT v1 \cite{li2022patch} & S & 32 &  8/8 & 71.56 & 4/8 & 65.57 \\
 & PSAQ-ViT v2 \cite{li2023psaq} & S & 32 & 8/8 & 72.17 & 4/8 & 68.61 \\
 & FQ-ViT \cite{lin2021fq} & R & 1000 & 8/8 & 71.61 & 4/8  & 66.91 \\
 & RepQ-ViT \cite{li2023repq} & R & 32 & 8/8 & 72.05 & 4/8  & 68.75 \\
 & \textbf{CLAMP-ViT (Ours)} & S & 32 & 8/8  & \textbf{\underline{72.17}} & 4/8 & \textbf{\underline{69.93}} \\
\Xhline{1\arrayrulewidth}
 \multirow{7}{*}{DeiT-S}& Baseline & - & - & 32/32 & 79.85 & 32/32 & 79.85 \\ \cline{2-8}
 & PSAQ-ViT v1 \cite{li2022patch} & S & 32 & 8/8 & 76.92 & 4/8 & 73.23 \\
 & PSAQ-ViT v2 \cite{li2023psaq} & S & 32 & 8/8 & \textbf{79.56} & 4/8 & 76.36 \\
  & PTQ4ViT \cite{yuan2022ptq4vit} & R & 32 & 8/8 & 79.47 & 4/8 & - \\
 & FQ-ViT \cite{lin2021fq} & R & 1000 & 8/8 & 79.17 & 4/8  & 76.93 \\
 & RepQ-ViT \cite{li2023repq} & R & 32 & 8/8 & 79.55 & 4/8  & 76.75 \\
 & \textbf{CLAMP-ViT (Ours)} & S & 32 & 8/8 & \underline{79.55} & 4/8 & \textbf{\underline{77.03}} \\
 \Xhline{1\arrayrulewidth}


   \multirow{6}{*}{Swin-T}& Baseline & - & - & 32/32 & 81.35 & 32/32 & 81.35 \\ \cline{2-8}
 & PSAQ-ViT v1 \cite{li2022patch} & S & 32 & 8/8 & 75.35 & 4/8 & 71.79 \\
 & PSAQ-ViT v2 \cite{li2023psaq} & S & 32 & 8/8 & 80.21 & 4/8 & 76.28 \\
 & FQ-ViT \cite{lin2021fq} & R & 1000 & 8/8 & \textbf{81.29} & 4/8  & \textbf{80.73} \\
 & RepQ-ViT \cite{li2023repq} & R & 32 & 8/8 & 81.28 & 4/8  & 80.51 \\
 & \textbf{CLAMP-ViT (Ours)} & S & 32 & 8/8 & \underline{81.17} & 4/8 & \underline{80.28} \\
 \Xhline{1\arrayrulewidth}
   \multirow{6}{*}{Swin-S}& Baseline & - & - & 32/32 & 83.20 & 32/32 & 83.20 \\ \cline{2-8}
 & PSAQ-ViT v1 \cite{li2022patch} & S & 32 & 8/8 & 76.64 & 4/8 & 75.14 \\
 & PSAQ-ViT v2 \cite{li2023psaq} & S & 32 & 8/8 & 82.13 & 4/8 & 78.86 \\
 & FQ-ViT \cite{lin2021fq} & R & 1000 & 8/8 & 82.13 & 4/8  & 81.67 \\
 & RepQ-ViT \cite{li2023repq} & R & 32 & 8/8 & 82.34 & 4/8  & 82.14\\
 & \textbf{CLAMP-ViT (Ours)} & S & 32 & 8/8 & \textbf{\underline{82.57}} & 4/8 & \textbf{\underline{82.51}} \\
 
 \Xhline{2\arrayrulewidth}
\end{tabular}
}
\vspace{-10pt}
\label{tab:classification}
\end{table}
\endgroup

\begingroup	
\begin{table}[t]\centering
 \caption{Mixed-Precision Quantization accuracy comparison on image classification tasks with ImageNet-1k testset. The values in \textbf{bold} indicate best performance overall.}
 \vspace{-10pt}
  \renewcommand*{\arraystretch}{1.0}
  \setlength\tabcolsep{1.9pt}
\resizebox{0.95\linewidth}{!}
{%
\begin{tabular}{cc|cc|cc|cc|cc}
\Xhline{2\arrayrulewidth}
& & \multicolumn{2}{c|}{DeiT-T} & \multicolumn{2}{c|}{DeiT-S} & \multicolumn{2}{c|}{Swin-T} & \multicolumn{2}{c}{Swin-S} \\
\Xhline{1\arrayrulewidth}
 Method & Data & W/A & Top-1 & W/A & Top-1 & W/A & Top-1 & W/A & Top-1  \\
\Xhline{2\arrayrulewidth}
 LRP-QViT \cite{ranjan2024lrp} & R & MP$_6$/MP$_6$ & 71.03 & MP$_6$/MP$_6$ & 79.03 & - & - & MP$_6$/MP$_6$ & \textbf{82.86} \\
 VT-PTQ \cite{liu2021post} & R & MP$_6$/MP$_6$ & 69.46 & MP$_6$/MP$_6$ & 75.10 & MP$_6$/MP$_6$ & 79.61 & MP$_6$/MP$_6$ & 78.43 \\
 \textbf{CLAMP-ViT (Ours)} & S & MP$_{4.9}$/MP$_{6.2}$ & \textbf{71.69} & MP$_{4.7}$/MP$_{5.9}$  & \textbf{79.43} & MP$_{5.5}$/MP$_{6.9}$ & \textbf{81.78} & MP$_{5.1}$/MP$_{6.3}$ & \textbf{82.86} \\
 \Xhline{2\arrayrulewidth}
\end{tabular}
}
\vspace{-4mm}
\label{tab:classification_mixed}
\end{table}
\endgroup

\textbf{Models and Datasets. } We evaluate CLAMP-ViT on various ViT model families (pre-trained FP models sourced from timm \cite{rw2019timm}) for image classification, object detection and semantic segmentation detailed as follows.

\indent
\textit{Image Classification. }We use ImageNet-1K \cite{deng2009imagenet} having 50K testset, with  DeiT-B/T/S \cite{touvron2021training}, Swin-T/S \cite{liu2021swin}, and ViT-B/S \cite{dosovitskiy2020image} to evaluate accuracy.

\indent
\textit{Object detection. }We use the COCO 2017 dataset \cite{lin2014microsoft} having approximately 20K test data. Following \cite{li2023psaq, lin2021fq, li2023repq}, we use the Cascade Mask R-CNN \cite{cai2018cascade} framework from MMdetection library \cite{chen2019mmdetection} with DeiT-S and Swin-S as the backbone.

\indent
\textit{Semantic Segmentation. }We use the ADE20K dataset \cite{zhou2019semantic} with 3K test data encompassing 150 categories with DeiT-S and Swin-S as the backbone. We adopt the UperNet framework \cite{xiao2018unified} in the MMsegmentation library \cite{mmseg2020} similar to \cite{li2023psaq}. 
\noindent
\textbf{Baselines. }
CLAMP-ViT is evaluated against SoTA PTQ (real data) and DFQ (synthetic data) methods for quantizing models from FP in various vision tasks. For image classification, it's compared with fixed-precision methods like PSAQ-ViT v1 \cite{li2022patch}, v2 \cite{li2023psaq}, FQ-ViT \cite{lin2021fq}, RepQ-ViT \cite{li2023repq}, and PTQ4ViT \cite{yuan2022ptq4vit}, and mixed-precision methods such as LRP-QViT \cite{ranjan2024lrp} and VT-PTQ \cite{liu2021post}. In object detection (Mask R-CNN), we use fixed-precision baselines FQ-ViT \cite{lin2021fq}, PSAQ-ViT v2 \cite{li2023psaq}, RepQ-ViT \cite{li2023repq}, and  LRP-QViT \cite{ranjan2024lrp} for mixed-precision comparison.
\begin{wraptable}{r}{0.46\textwidth}
\vspace{-12mm}
	\tiny
		\caption{Hyperparameters list.}
		\label{tab:parameters}
		\centering
		      \begin{tabular}{ccc}
            \Xhline{2\arrayrulewidth}
            Parameter & Description & Value \\
            \Xhline{2\arrayrulewidth}
            $G$ & Generation Iterations & 500 \\
            $\mathcal{N}$ & Neighborhood Size & $3 \times 3$ \\
            $n$ & \# Positive Patches & 4 \\
            $\mathcal{P}$ & Passes & 10 \\
            $\mathcal{C}$ & Cycles & 6 \\
            $\mathcal{K}$ & \# Initial Candidates & 15 \\
            $\mathcal{B}$ & Batch Size & 32 \\
            $b$ & Bit-width & [2,8] \\
            \Xhline{2\arrayrulewidth}
        \end{tabular}
		\vspace{-7mm}
\end{wraptable}
For semantic segmentation (UperNet), PSAQ-ViT v2 \cite{li2023psaq} serves as the baseline. Evol-Q is excluded from the main comparison as it does not fully quantize a model starting from FP, and is presented  for ablations in the supplementary.

\noindent
\textbf{Experimental Setup. }The CLAMP-ViT framework is implemented in PyTorch, and evaluated on a single NVIDIA Titan GPU. It features multiple hyperparameters, detailed in \cref{tab:parameters}.

\subsection{Quantization Results for Image Classification}
\label{sec:classification}
\begin{wraptable}{r}{0.46\textwidth}
\vspace{-12mm}
	\tiny
		\caption{Comparison of quantized DeiT-S size (MB) and BOPS (G) \cite{baskin2021uniq}.}
		\label{tab:size}
		\centering
		\begin{tabular}{cccc}
\Xhline{2\arrayrulewidth}
Method & W/A & BOPS & Size \\
\Xhline{2\arrayrulewidth}
Baseline & 32/32 & 4710 & 88 \\
\Xhline{1\arrayrulewidth}
PSAQ-ViT v2 & 4/8 & 294 & 22 \\
\textbf{CLAMP-ViT (Ours) }& MP$_{4.7}$/MP$_{5.9}$ & \textbf{267} & \textbf{20} \\
 \Xhline{2\arrayrulewidth}
\end{tabular}
\vspace{-8mm}
\end{wraptable}

As highlighted in \cref{sec:pipeline}, CLAMP-ViT requires a batch $\mathcal{B}$ of task-specific targets $T_{G_{\mathcal{B}}}$. For image classification on the ImageNet-1K, we create $T_{G_{\mathcal{B}}} \in \mathcal{R}^{\mathcal{B} \times 1000}$, where the class-wise probabilities are randomly determined and assigned. We discuss and compare the performance of CLAMP-ViT for two settings, fixed-precision (\cref{tab:classification}) and mixed-precision (\cref{tab:classification_mixed}). 
In specific, as shown in \cref{tab:classification} CLAMP-ViT consistently provides similar or better accuracy at W8/A8, while for \textbf{lower precision W4/A8 CLAMP-ViT shows significant performance boost over all the existing alternatives}.  We yield $\sim \textbf{2.2} \%$ and $\sim \textbf{1} \%$ average accuracy improvement compared to  DFQ methods \cite{li2023psaq, li2022patch} and data-driven methods, respectively. The superior performance of CLAMP-ViT can be attributed to the cyclically adaptive data-generation process, which ensures the generated data matches the requirements and representational capabilities of the quantized model and effective traversal of the search space through evolutionary search and contrastive learning. Whereas, PSAQ-ViT v2 \cite{li2023psaq}, generates increasingly difficult samples which is less beneficial for aggressive 4-bit quantization. Surprisingly, PSAQ-ViT v1 \cite{li2022patch} achieves poor accuracy of $25.34\%$ (W4/A8) on ViT-B despite achieving reasonable accuracy on other ViTs. This result potentially supports our initial intuition that  PSAQ-ViT \cite{li2023psaq, li2022patch} does not consider the informativeness of the generated data to the quantization process.

\begingroup	
\begin{table}[t]\centering
 \caption{Mixed and fixed-precision quantization performance comparison against SoTA techniques for object detection on COCO 2017. The values in \textbf{bold} and \underline{underline} signifies, the best performance overall, and with synthetic data, respectively.  }
 \vspace{-10pt}
  \renewcommand*{\arraystretch}{1.0}
  \setlength\tabcolsep{1.9pt}
\resizebox{0.7\linewidth}{!}
{%
\begin{tabular}{cc|ccc|ccc}
\Xhline{2\arrayrulewidth}
& & \multicolumn{3}{c|}{DeiT-S} & \multicolumn{3}{c}{Swin-S}  \\
\Xhline{1\arrayrulewidth}
 Method & Data & W/A & AP$^{box}$ & AP$^{mask}$ & W/A & AP$^{box}$ & AP$^{mask}$  \\
\Xhline{2\arrayrulewidth}
Baseline & - & 32/32 & 48.0 & 41.4 & 32/32 & 51.8 & 44.7 \\
  \Xhline{1\arrayrulewidth}

 FQ-ViT \cite{lin2021fq} & R & 8/8 & 47.4 & 40.9 & 8/8 & 50.8 & 44.1 \\
 PSAQ-ViT v2 \cite{li2023psaq} & S & 8/8 & 47.3 & 40.8 & 8/8 & 50.9 & 44.1 \\
 RepQ-ViT \cite{li2023repq} & R & 8/8 & \textbf{47.9} & \textbf{41.1} & 8/8 & \textbf{51.6} & \textbf{44.6} \\
  \textbf{CLAMP-ViT (Ours)} & S & 8/8 & \underline{\textbf{47.9}}  & \underline{\textbf{41.1}} & 8/8 & \underline{51.4}  & \underline{\textbf{44.6}} \\
 \Xhline{1\arrayrulewidth}
  FQ-ViT \cite{lin2021fq} & R & 4/8 & 45.1 & 40.2 & 4/8 & 48.2 & 41.3 \\
  PSAQ-ViT v2 \cite{li2023psaq} & S & 4/8 & 44.8 & 38.8 & 4/8 & 47.9 & 41.4 \\
  RepQ-ViT \cite{li2023repq} & R & 4/8 & \textbf{45.6} & \textbf{39.5} & 4/8 & \textbf{49.2} & \textbf{42.8} \\
 \textbf{CLAMP-ViT (Ours)} & S & 4/8 & \underline{45.4} & \underline{38.9} & 4/8 & \underline{48.5} & \underline{42.2} \\
 \Xhline{1\arrayrulewidth}

 LRP-QViT \cite{ranjan2024lrp} & R & - & - & - & MP$_6$/MP$_6$  & 51.4 & \textbf{44.6} \\
 \textbf{CLAMP-ViT (Ours)} & S & MP$_{5.5}$/MP$_{6.8}$ & \underline{\textbf{47.9}} & \underline{\textbf{41.0}} & MP$_{5.1}$/MP$_{6.4}$ & \underline{\textbf{51.7}} & \underline{\textbf{44.6}}\\
 \Xhline{2\arrayrulewidth}
\end{tabular}
}
\vspace{-10pt}
\label{tab:object}
\end{table}
\endgroup
Evident from \cref{tab:classification_mixed}, CLAMP-ViT consistently outperforms all baselines across models for the mixed-precision setting, maintaining accuracy close to the FP baseline despite having a significantly lower average W/A (for mixed-precision W/A is calculated by averaging bit-widths over all the layers for weights and activations). Furthermore, in \cref{tab:size}, we report the reduction in model size and \textit{bit-operations-per-second} (BOPS) \cite{baskin2021uniq} of mixed-precision quantized DeiT-S compared with W4/A8 PSAQ-ViT v2.
CLAMP-ViT achieves \textbf{$\sim 10\%$ lower model size in MB and BOPS when compared to W4/A8 PSAQ-ViT v2 while yielding $\textbf{3.07\%}$ improved accuracy}. 



\begingroup	
\begin{table}[t]\centering
 \caption{Mixed-precision quantization performance comparison against PSAQ-ViT v2 for semantic segmentation. The values in \textbf{bold} indicate best performance overall.}
 \vspace{-6pt}
  \renewcommand*{\arraystretch}{1.0}
  \setlength\tabcolsep{1.9pt}
\resizebox{0.6\linewidth}{!}
{%
\begin{tabular}{cc|cc|cc}
\Xhline{2\arrayrulewidth}
& & \multicolumn{2}{c|}{DeiT-S} & \multicolumn{2}{c}{Swin-S}  \\
\Xhline{1\arrayrulewidth}
 Method & Data & W/A & mIoU & W/A & mIoU  \\
\Xhline{2\arrayrulewidth}
Baseline & - & 32/32 & 44.0 & 32/32 & 49.3 \\
  \Xhline{1\arrayrulewidth}
  PSAQ-ViT v2 \cite{li2023psaq} & S & 4/8 & 39.9 & 4/8 & 44.6 \\
  \textbf{CLAMP-ViT (Ours)} & S & MP$_{4.8}$/MP$_{6.2}$ & \textbf{42.4} & MP$_{5.1}$/MP$_{6.4}$ & \textbf{45.9} \\
 \Xhline{2\arrayrulewidth}
\end{tabular}
}
\vspace{-10pt}
\label{tab:semantic}
\end{table}
\endgroup
\subsection{Quantization Results for Object Detection}
\label{sec:object}
The target for object detection is $T_{G_{\mathcal{B}}} \in \mathcal{R}^{\mathcal{B} \times bb \times 5}$ where $bb$ is the number of bounding boxes in the image that is randomly selected from the integer set $[1,3]$. $T_{G_{\mathcal{B}}}[\mathcal{B}, :, 0]$ corresponds to the bounding box category and $T_{G_{\mathcal{B}}}[\mathcal{B}, :, 1:4]$ is the bounding box coordinates $x,y,w,h$ \cite{huang2018introduction}. \cref{tab:object} presents the fixed- and mixed-precision performance of CLAMP-ViT with respect to the baselines. Across different settings and models, CLAMP-ViT \textbf{consistently outperforms DFQ method PSAQ-ViT v2 \cite{li2023psaq} by 0.6 box AP and 0.4 mask AP on average} while closely matching performance to the SoTA data-driven method, RepQ-ViT \cite{li2023repq}. Similar to \cref{sec:classification}, we observe improved performance with mixed-precision quantization, achieving near FP baseline performance. The average W/A for mixed-precision quantization for object detection is found to be higher than that for image classification due to the higher complexity of the task demanding larger bit-widths to maintain accuracy.

\subsection{Quantization Results for Semantic Segmentation}
\label{sec:semantic}
The target $T_{G_{\mathcal{B}}}$ for this task is a pixel-wise classification map of the same size as $X_{\mathcal{B}}$ i.e, $T_{G_{\mathcal{B}}} \in \mathcal{R}^{\mathcal{B} \times 150 \times H \times W}$. In \cref{tab:semantic}, we show the quantization performance comparison, where CLAMP-ViT achieves average weight bit-width close to 4-bit while also significantly reducing activation bit-width to $\sim6$-bits. \textbf{At a higher compression ratio of $\textbf{15}\%$, CLAMP-ViT achieves $\textbf{1.5}$ mIoU improvement on average over PSAQ-ViT v2}.

\begin{figure}[!t]
  \centering \includegraphics[width = 0.8\columnwidth, keepaspectratio]{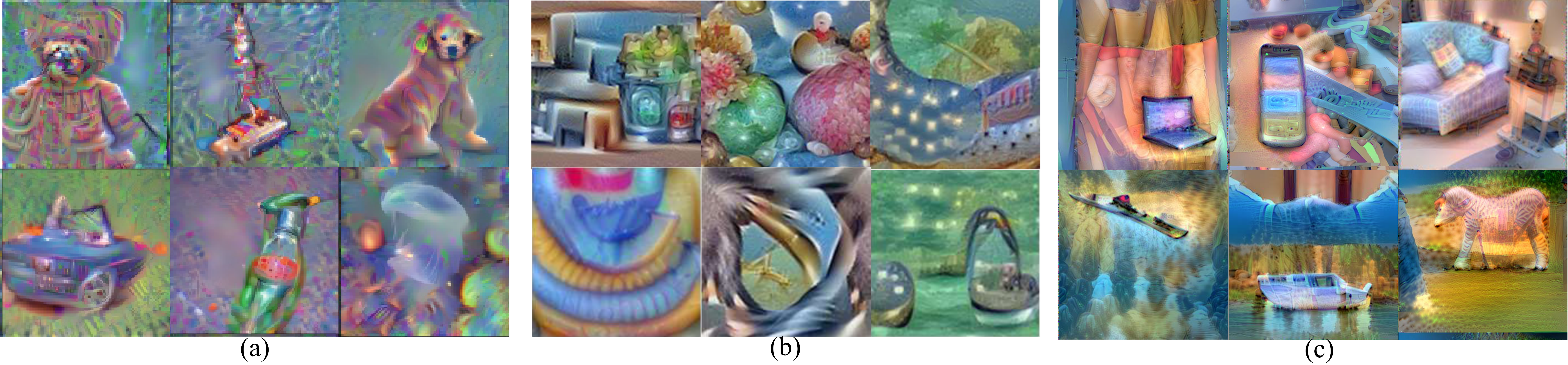}
  \caption{Comparison of synthetic data generated by (a) PSAQ-ViT v1 \cite{li2022patch}, (b) PSAQ-ViT v2 \cite{li2023psaq} and (c) \textbf{CLAMP-ViT (Ours)}. CLAMP-ViT generates detailed objects within contextually suitable backgrounds, boosting realism and informativeness.}  
  \vspace{-5mm}
  \label{fig:sample}
\end{figure}

\subsection{Analysis of Generated Samples}
\label{sec:sample_generation}
\cref{fig:sample} visualizes generated samples from PSAQ-ViT v1 \cite{li2022patch}, v2 \cite{li2023psaq}, and CLAMP-ViT (after $1^{st}$ round of stage 1 execution). PSAQ-ViT v1 (\cref{fig:sample}(a)) creates images with clear class-specific foregrounds but with overly simplistic and uniform backgrounds, resulting in a lack of realism potentially affecting model accuracy. PSAQ-ViT v2 (\cref{fig:sample}(b)) introduces more complex details but fails to convey meaningful semantic information, generating images with intricate but semantically vague structures due to its unguided, difficulty-increasing data-generation strategy. 
In contrast, CLAMP-ViT (\cref{fig:sample}(c)) excels by synthesizing data that mirrors real-world imagery, showcasing a sophisticated understanding of semantic relationships between patches.It ensures objects are detailed and in contextually fitting backgrounds, boosting realism and informativeness. For example, CLAMP-ViT places boats on water and zebras in grasslands (\cref{fig:sample}(c), row 2), showing its capability for creating semantically relevant and visually consistent synthetic data. We believe our patch semantics exploration with a contrastive objective, makes image generation informative that mimic real-world scenes.


\subsection{Ablations and Discussions}
\label{sec:ablation}
\begin{figure}[t]
  \centering \includegraphics[width = \columnwidth, keepaspectratio]{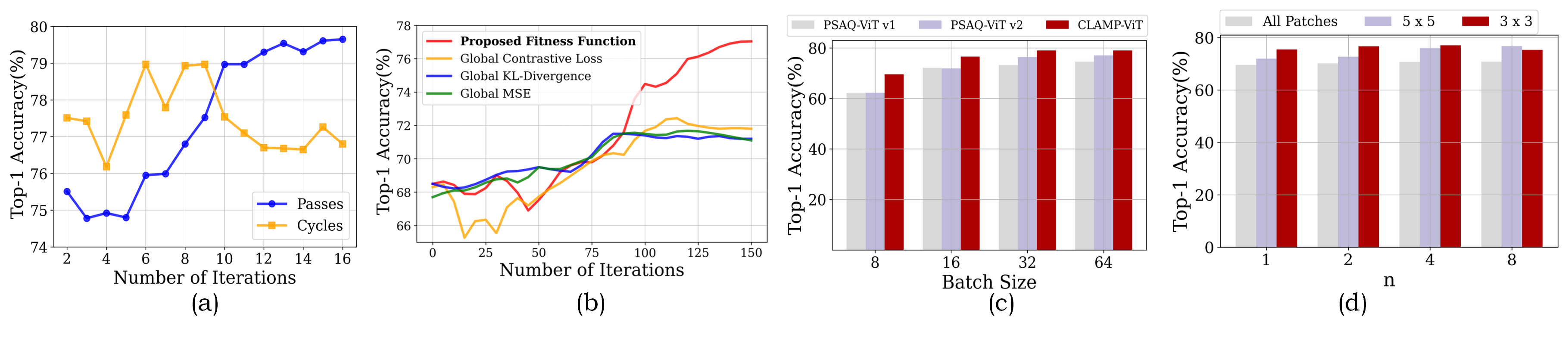}
  \vspace{-6mm}
  \caption{CLAMP-ViT ablations for (a) Selecting evolutionary search parameters, (b) Mixed-precision quantization accuracy with different fitness functions, (c) Effect of batch size $\mathcal{B}$ and (d) Effect of neighborhood size $\mathcal{N}$ and top-$n$ positive patches.}
  \vspace{-8mm}
  \label{fig:ablation_figs}
\end{figure}

\textbf{Evolutionary Search Parameters. } In \cref{fig:ablation_figs}(a), we detail an experiment to determine the ideal number of passes $\mathcal{P}$ and cycles $\mathcal{C}$ for the evolutionary search process by studying the variation in Top-1 accuracy of DeiT-S with different passes and cycles keeping the other fixed at their optimal value (\cref{tab:parameters}). It is evident that a cycle count of $\mathcal{C}$=6 is optimal, as accuracy tends to decline with more cycles. Conversely, passes show a modest yet consistent improvement beyond 10, but due to the substantial rise in computational complexity, $\mathcal{P}$=10 is deemed the most suitable choice.

\noindent
\textbf{Effect of Batch Size $\mathcal{B}$. }We also show the accuracy comparison with different batch sizes ranging from 8 to 64 in \cref{fig:ablation_figs}(c). It is apparent that there is minimal increase in accuracy beyond 32 for CLAMP-ViT, justfiying the choice of batch size. Interestingly, CLAMP-ViT achieves a decent accuracy of $\sim 70\%$ even with a batch size of 8, while PSAQ-ViT v1 and v2 achieve only close to $60\%$.

\noindent
\textbf{Effect of Neighborhood Size $\mathcal{N}$ and Top-$n$ Patches. } We explore $3 \times 3$, $5 \times 5$, and all patches for neighborhood sizes and $n \in [1,4]$ for DeiT-S as shown in \cref{fig:ablation_figs}(d). A $3 \times 3$ neighborhood with top-4 patches as positives yields the highest accuracy while being computationally least demanding. In the $5 \times 5$ scenario, as the balance between positive and negative samples improves with increasing $n$, accuracy rises but at higher computational cost. Employing all patches is computationally unviable and leads to the lowest accuracy for small $n$ values due to a positive-negative imbalance \cite{wang2023positive}. Moreover, identifying distant positive patches from the anchor neglects significant semantic patch relationships.

\begin{wraptable}{r}{0.46\textwidth}
\vspace{-12mm}
	\tiny
		\caption{Impact of different loss components in synthetic data generation.}
		\label{tab:loss_func}
		\centering
		\begin{tabular}{ccccc}
            \Xhline{2\arrayrulewidth}
            $\mathcal{L}^{PSE}$ & $\mathcal{L}^{C_1}$ & $\mathcal{L}^{O}$ & W/A & Top-1 Accuracy  \\
            \Xhline{2\arrayrulewidth}
             - & - & - & 32/32 & 79.85 \\ \cline{1-5}
             \cmark & \xmark & \xmark & 4/8 & 74.18 \\
             \xmark & \xmark & \cmark & 4/8 & 33.93 \\
             \cmark & \xmark & \cmark & 4/8 & 76.47 \\
             \xmark & \cmark & \xmark & 4/8 & 76.59 \\ 
             \xmark & \cmark & \cmark & 4/8 & \textbf{77.03} \\
             \cmark & \cmark & \cmark & 4/8 & 76.98 \\
            \Xhline{2\arrayrulewidth}
        \end{tabular}
		\vspace{-7mm}
\end{wraptable}

\noindent
\textbf{Choice of Objective Function. }
The study in \cref{tab:loss_func} examines how different loss function components of $\mathcal{L}^{SG}$ i.e., $\mathcal{L}^{C_1}$ and $\mathcal{L}^O$ affect synthetic data generation effectiveness and its impact on top-1 accuracy of W4/A8 quantization of DeiT-S.We chose fixed-precision quantization to avoid any bias from mixed-precision quantization, which might typically favor higher bitwidths to lessen accuracy loss due to low-quality images. Results show that a linear mix of $\mathcal{L}^{C_1}$ and $\mathcal{L}^O$ ($\mathcal{L}^{SG}$) achieves highest accuracy, while using $\mathcal{L}^O$ alone leads to the lowest, indicating its limited utility in leveraging ViTs for synthetic data. In \cref{tab:loss_func} $\mathcal{L}^{PSE}$ corresponds to the patch similarity metric employed in \cite{li2023psaq}. 
While combining $\mathcal{L}^{PSE}$ with $\mathcal{L}^O$ \cite{li2023psaq} does offer a moderate accuracy boost, it falls short of the performance with $\mathcal{L}^{SG}$, due to inherent limitations of the patch similarity metric highlighted in \cref{sec:introduction}. Furthermore, using all three loss functions simultaneously closely matches the performance of $\mathcal{L}^{SG}$ further demonstrating that our proposed $\mathcal{L}^{C_1}$ has the major contribution towards final accuracy.

For mixed-precision quantization of DeiT-S, the fitness function's accuracy ($\mathcal{L}^F = \mathcal{L}^{C_2} +\mathcal{L}^O$) is compared against global contrastive loss \cite{frumkin2023jumping}, MSE, and KL-divergence in \cref{fig:ablation_figs}(b). The accuracy analysis shows MSE and KL-divergence tend to overfit to synthetic data, evidenced by plateauing accuracy. Meanwhile, global contrastive loss initially matches but then accuracy gap widens from CLAMP-ViT's performance which is due to premature convergence, implying that quantifying intermediate layer distributional divergence is crucial to find optimal quantization parameters. 

\begin{wraptable}{r}{0.46\textwidth}
\vspace{-12mm}
  \tiny
		\caption{Effect of adaptability on quantized model accuracy.}
		\label{table:cyclic}
		\centering
		\begin{tabular}{ccc}
\Xhline{2\arrayrulewidth}
Adaptivity & W/A & Top-1 Accuracy \\
\Xhline{2\arrayrulewidth}
\xmark & MP$_{4.9}$/MP$_{6.1}$ & 78.06  \\
\cmark & MP$_{4.7}$/MP$_{5.9}$ & \textbf{78.97}  \\
 \Xhline{2\arrayrulewidth}
\end{tabular}
		\vspace{-7mm}
\end{wraptable}
\noindent
\textbf{Effect of Adaptivity. }We investigate the effectiveness of the cyclic evolution every $\mathcal{C}/2$ iterations to ensure the data generation adapts to the requirements of the quantized model in \cref{table:cyclic}. It can be observed that with adaptivity, we are not only able to achieve lower average W/A but also have improved accuracy.
\begin{wrapfigure}{R}{0.40\textwidth}
\vspace{-8mm}
  \centering \includegraphics[width = 0.3\textwidth, keepaspectratio]{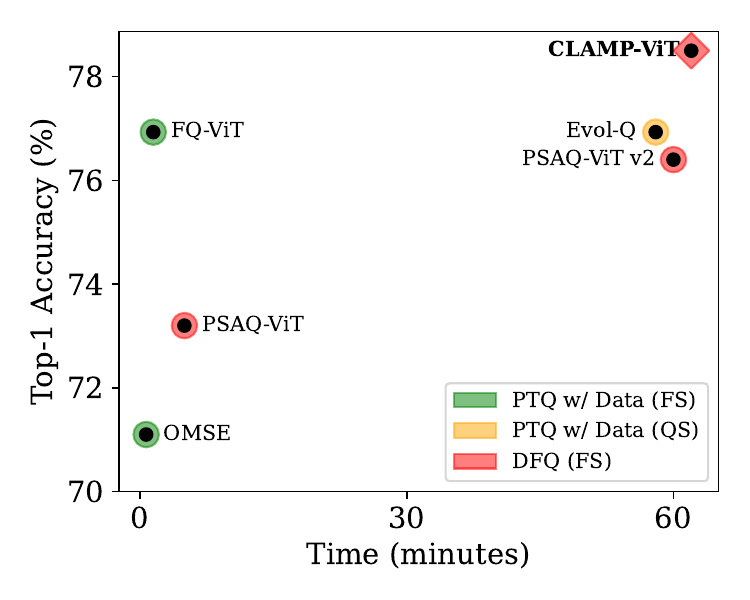}
  \vspace{-4mm}
  \caption{Runtime of different PTQ and DFQ methods where, \textbf{FS:} FP model as starting point. \textbf{QS:} Pre-quantized model as starting point.}
  \vspace{-8mm}
  \label{fig:runtime}
\end{wrapfigure}

\noindent
\textbf{Effect of Informativeness. }
To show CLAMP-ViT's effectiveness in generating informative data, we compare misprediction rates in PSAQ-ViT v1, v2, and CLAMP-ViT during quantization. PSAQ-ViT has higher misprediction rates (34\% for v1, 41\% for v2) than CLAMP-ViT's 22\%, indicating PSAQ-ViT's data is less informative, leading to erratic predictions, sub-optimal performance.

\noindent
\textbf{Limitation Discussion: Runtime Comparison. }In \cref{fig:runtime}, we show the runtime and top-1 accuracy of different techniques on an NVIDIA Titan GPU for DeiT-S. CLAMP-ViT (mixed-precision) achieves significantly higher accuracy compared to the other methods (fixed-precision) with only a minimal increase in runtime ($5\% \uparrow$) compared to the best DFQ method. Note, Evol-Q takes similar time to calibrate a pre-quantized model on real data.  


\section{Conclusions}
This paper presents CLAMP-ViT, a novel mixed-precision DFQ technique using cyclic adaptation and contrastive learning. It employs patch-level contrastive learning that leverages properties of the MHSA modules for data generation. A local contrastive objective and layer-wise evolutionary search identify optimal quantization parameters while ensuring a smooth loss landscape. Experiments across CV tasks show superior performance of CLAMP-ViT, achieving up to 3\% top-1 accuracy for classification, 0.6 mAP for detection, and 1.5 mIoU for segmentation. Future work aims to focus on extending its application to a wider range of architectures, like VLMs. While this study focuses on a useful impact and beneficial application of synthetic data generation for optimized and carbon efficient models for deployment, it is important to also be cognizant of the potential adverse effects of synthetic data such as deepfakes or racial biases.   

\section*{Acknowledgements}
This work was supported in part by CoCoSys, one of the seven centers in JUMP 2.0, a Semiconductor Research Corporation (SRC) program sponsored by DARPA.


%
%
\clearpage
\chapter*{Supplementary Material}
\addcontentsline{toc}{chapter}{Supplementary Material}
\renewcommand\thesection{\Alph{section}}
\renewcommand\thesubsection{\thesection.\arabic{subsection}}
\setcounter{section}{0}
\newcommand{\ccomment}[2][black]{\textcolor{#1}{\emph{\# \textbf{#2}}}}
\SetAlFnt{\scriptsize}
\vspace{-40pt}
\begin{algorithm}
    \SetKwInOut{Input}{Input}
    \SetKwInOut{Output}{Output}
    \caption{CLAMP-ViT Pipeline}
    \Input{$\mathcal{B}$ randomly produced Gaussian Images $X_{{\mathcal{B}}}$, $\mathcal{B}$ task-specific targets $T_{G_{\mathcal{B}}}$, Pre-Trained FP ViT, Quantized Model $Q$ initialized to a best search candidate within population, Initial candidate population of $\mathcal{K}$ tuples, \{$(\Lambda_0, \mathcal{L}^F_0)$, ... , $(\Lambda_{k-1}, \mathcal{L}^F_{k-1})$ \} }

    \Output{Fully quantized ViT $Q$}
\ccomment[orange]{Stage 1: Sample Generation (FP and Q remain fixed)}\

    \textbf{stage\_1: }\For{G iterations}{
    Input $X_{{\mathcal{B}}}$ into FP and Q;
    
    Obtain $\mathcal{L}^{C_1}$ according to Eq.(4);
    
    Obtain $\mathcal{L}^O$ between logits output of FP and Q with $T_{G_{\mathcal{B}}}$;
    
    Combine the losses to obtain sample generation loss $\mathcal{L}^{SG} = \mathcal{L}^{C_1} + \mathcal{L}^O$ ;
    
    Update $X$ by minimizing $\mathcal{L}^{SG}$;
  }

\ccomment[orange]{Stage 2: Quantization ($X_{\mathcal{B}}$ and FP remain fixed.)}\

    \textbf{stage\_2: }\For{$\mathcal{P}$ passes}{
    \For{each transformer layer i}{
    \For{$\mathcal{C}$ cycles}{
    Input $X_\mathcal{B}$ into FP and Q;
    
    Select top two candidates from population with best $\mathcal{L}^F$ as parents $p_1, p_2$;

    \ccomment[pink]{Regeneration}\

    Perform mutation and crossover on $p_1, p_2$ for the $i^{th}$ transformer layer to generate child parameters $C0$ (Eq.(6), Eq.(7)).

    Perform diversity-promoting selection to generate additional diverse child candidates $C1, .... C5$.

    \ccomment[pink]{Evaluation and Population Update}\

    Obtain $\mathcal{L}^F$ for each child candidate;

    Add C0 and best diverse child along with corresponding $\mathcal{L}^F$ as a tuple to population;

    Pop the worst two candidates from the population;

    \ccomment[pink]{Activation Quantization}\
    
    Estimate quantization parameters of output activations;
    
    \ccomment[pink]{Cyclic Adaptation}\
    
    \If{cycles $== \mathcal{C}/2$}{

    \textbf{goto} stage\_1;
    
    Update $X_\mathcal{B}$ for \textit{G}/2 iterations;
    }
    }
    }    
    }

 \label{algo:clamp_vit}   
\end{algorithm}

\section{CLAMP-ViT Algorithm}
In \cref{algo:clamp_vit}, we summarize the whole pipeline of the proposed CLAMP-ViT framework as an aid to better understand the discussions in our main paper.

\begingroup	
\begin{table}[t]\centering
 \caption{Fixed-precision quantization accuracy comparison with SoTA on image classification tasks with ImageNet-1k testset for ViT-S and DeiT-B.}
 \vspace{-10pt}
  \renewcommand*{\arraystretch}{1.0}
  \setlength\tabcolsep{1.9pt}
\resizebox{0.7\linewidth}{!}
{%
\begin{tabular}{cccc|cc|cc}
\Xhline{2\arrayrulewidth}
Model & Method & Data & \#Images &  W/A & Top-1 & W/A & Top-1 \\
\Xhline{2\arrayrulewidth}

  \multirow{6}{*}{ViT-L}& Baseline & - & - & 32/32 & 81.39 & 32/32 & 81.39 \\ \cline{2-8}
 & PSAQ-ViT v1 \cite{li2022patch} & S & 32 & 8/8 & 31.45 & 4/8 & 20.84 \\
  & FQ-ViT \cite{lin2021fq} & R & 1000 & 8/8 & 79.68 & 4/8  & 75.49 \\
 & RepQ-ViT \cite{li2023repq} & R & 32 &  8/8 & \textbf{81.19} & 4/8  & 79.48 \\
 & \textbf{CLAMP-ViT (Ours)} & S & 32 & 8/8 & \underline{81.15} & 4/8 & \textbf{\underline{80.06}} \\
 \Xhline{1\arrayrulewidth}

  \multirow{7}{*}{DeiT-B}& Baseline & - & - & 32/32 & 81.85 & 32/32 & 81.85 \\ \cline{2-8}
 & PSAQ-ViT v1 \cite{li2022patch} & S & 32 & 8/8 & 79.10 & 4/8 & 77.05 \\
 & PSAQ-ViT v2 \cite{li2023psaq} & S & 32 & 8/8 & 81.52 & 4/8 & 79.49 \\
  & PTQ4ViT \cite{yuan2022ptq4vit} & R & 32 & 8/8 & 81.48 & 4/8 & 64.39 \\
 & FQ-ViT \cite{lin2021fq} & R & 1000 & 8/8 & 81.20 & 4/8  & 79.99 \\
 & RepQ-ViT \cite{li2023repq} & R & 32 &  8/8 & 81.45 & 4/8  & 80.12 \\
 & \textbf{CLAMP-ViT (Ours)} & S & 32 & 8/8 & \textbf{\underline{81.77}} & 4/8 & \textbf{\underline{80.93}} \\
 \Xhline{1\arrayrulewidth}

 \Xhline{2\arrayrulewidth}
\end{tabular}
}
\vspace{-1pt}
\label{tab:classification_sup}
\end{table}
\endgroup

\begin{table}[t]
\vspace{-3mm}
	\begin{minipage}{0.4\linewidth}
	    \vspace{-4mm}
		\caption{Quantized W4/A4 DeiT-S top-1 acc. on ImageNet testset.}
		\label{tab:low_precision}
		\centering
		\begin{tabular}{cc|c|c}
		\Xhline{1.5\arrayrulewidth}
             Method & Data & W/A &  Acc. $\%$  \\
            \Xhline{1.5\arrayrulewidth}
            Baseline & - & 32/32 & 79.85 \\
              \Xhline{1\arrayrulewidth}
              FQ-ViT \cite{lin2021fq} & R & 4/4 & 0.10 \\
              PTQ4ViT \cite{yuan2022ptq4vit} & R & 4/4 & 34.08 \\
              PSAQ-ViT v2 \cite{li2023psaq} & S & 4/4 & 57.97 \\
              \textbf{CLAMP-ViT} & S & 4/4 & \textbf{69.01} \\
             \Xhline{1.5\arrayrulewidth}
            \end{tabular}
            
	\end{minipage}\hfill
	\begin{minipage}{0.4\linewidth}
		\centering
		\includegraphics[width=\textwidth]{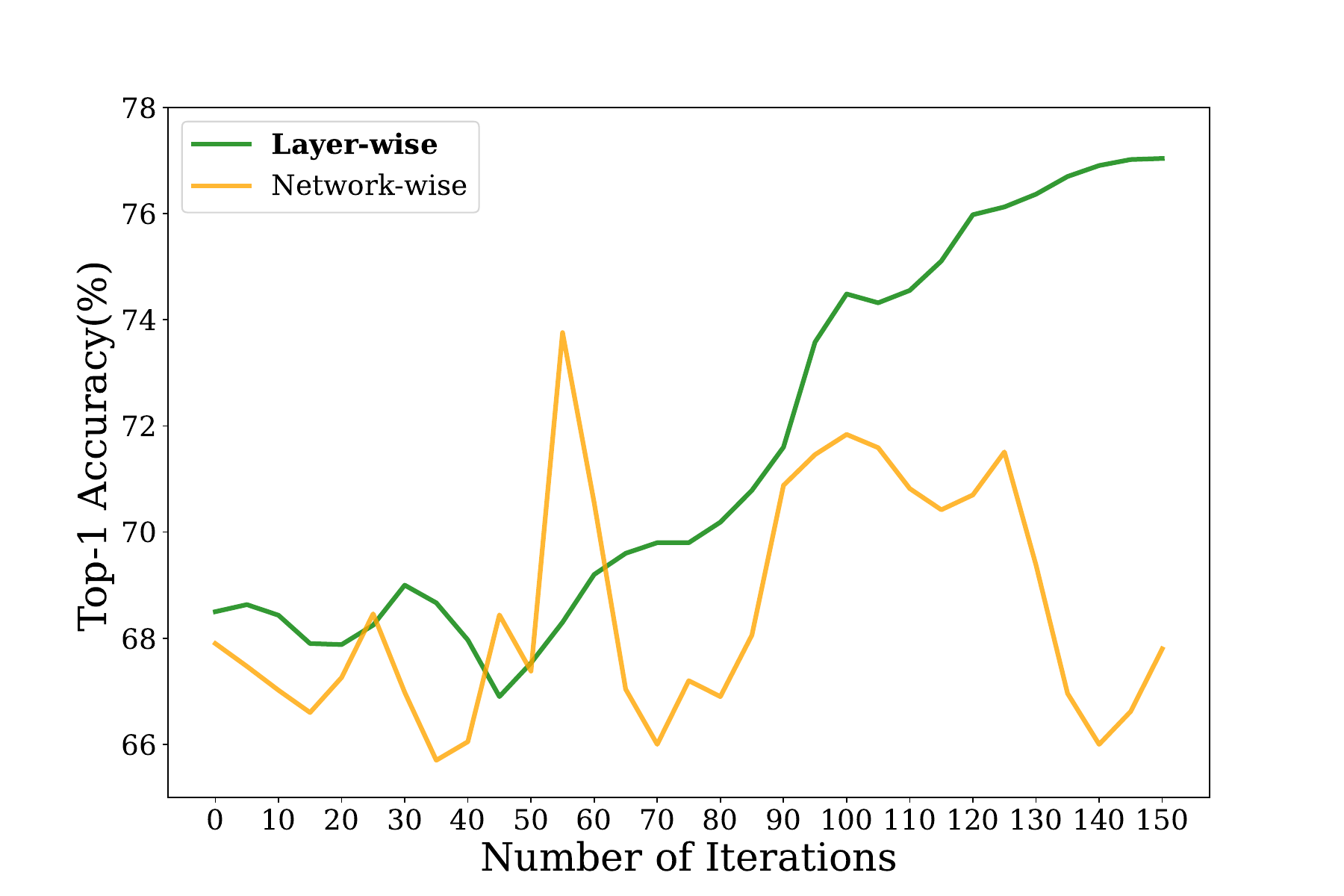}
            \vspace{-3mm}
		\captionof{figure}{Accuracy vs. iteration with perturbation to all (network-wise)  and layer-wise parameters.}
	\label{fig:iteration}
	\end{minipage}
	\vspace{-1mm}
\end{table}

\begingroup	
\begin{table}[t]\centering
 \caption{Comparison of quantized DeiT-S model size (MB) of CLAMP-ViT and PSAQ-ViT v2.}
 \vspace{-2pt}
  \renewcommand*{\arraystretch}{1.0}
  \setlength\tabcolsep{1.9pt}
\resizebox{0.7\linewidth}{!}
{%
\begin{tabular}{cccc}
\Xhline{2\arrayrulewidth}
Task & Method & W/A & Size \\
\Xhline{1\arrayrulewidth}
\multirow{3}{*}{Object Detection} & Baseline & 32/32 & 320 \\ \cline{2-4}
 & PSAQ-ViT v2 \cite{li2023psaq} & 4/8 & 40 \\
 & \textbf{CLAMP-ViT (Ours)} & MP$_{5.5}$/MP$_{6.8}$ & \textbf{39} \\
 \Xhline{1\arrayrulewidth}
\multirow{3}{*}{Semantic Segmentation} & Baseline & 32/32 & 208 \\ \cline{2-4}
 & PSAQ-ViT v2 \cite{li2023psaq} & 4/8 & 26 \\
 & \textbf{CLAMP-ViT (Ours)} & MP$_{4.8}$/MP$_{6.2}$ & \textbf{21} \\
\Xhline{2\arrayrulewidth}

\end{tabular}
}
\vspace{-1pt}
\label{tab:sizes}
\end{table}
\endgroup

\section{Additional Experiments}
\subsection{Quantization Results for Image Classification}
In \cref{tab:classification_sup}, we show performance comparison of CLAMP-ViT with the baselines for additional models-ViT-L and DeiT-S. CLAMP-ViT outperforms the baselines operating on real data by upto $1\%$ and DFQ methods by up to $60\%$. Similar to the results on ViT-B in the main paper, PSAQ-ViT v1 \cite{li2022patch} achieves poor accuracy for different bitwidth quantizations for ViT-L.

Additionally we also demonstrate low precision quantization results of weights and activations (W4A4) in \cref{tab:low_precision}. Evident from the table, CLAMP-ViT remains robust in the face of extreme quantization and outperforms DFQ and PTQ methods due to its cyclically adaptive strategy. This also demonstrates the importance of having the synthetic data adapt to the requirements of the quantization process. 

\subsection{Quantized Model Size Comparison}
We further evaluate the reduction in model sizes after mixed-precision quantization for object detection and semantic segmentation to further study and conclusively demonstrate that CLAMP-ViT mixed-precision quantization constanly results in lower model size than fixed-precision quantization. We showcase our findings in \cref{tab:sizes}, and we find that similar to the classification scenario highlighted in our main paper, CLAMP-ViT achieves upto $20\%$ lower quantized model size than PSAQ-ViT v2.

\begingroup	
\begin{table}[t]\centering
 \caption{Accuracy comparison with Evol-Q \cite{frumkin2023jumping} for image classification. Here, $\Delta_{Acc}^{Avg}$ repesents the average accuracy difference from that with Evol-Q, a +ve value identifies CLAMP-ViT to yield better average accuracy.}
 \vspace{-10pt}
  \renewcommand*{\arraystretch}{1.0}
  \setlength\tabcolsep{1.9pt}
\resizebox{\linewidth}{!}
{%
\begin{tabular}{cc|cc|cc|cc|c}
\Xhline{2\arrayrulewidth}
& & \multicolumn{2}{c|}{DeiT-T} & \multicolumn{2}{c}{DeiT-S} & \multicolumn{2}{c|}{Swin-S} &  \\
\Xhline{1\arrayrulewidth}
 Method & Data & W/A & Top-1 & W/A & Top-1 & Top-1 & W/A & $\Delta_{Acc}^{Avg}$ \\
\Xhline{2\arrayrulewidth}
Baseline & - & 32/32 & 72.21 & 32/32 & 79.85 & 32/32 & 83.20 & - \\
\Xhline{1\arrayrulewidth}
 Evol-Q & R & 8/8 & 71.63 & 8/8 & \textbf{79.57} & 8/8 & \textbf{82.98} & N/A \\
 \textbf{CLAMP-ViT (Ours)} & S & 8/8 & \textbf{72.17} & 8/8 & 79.55 & 8/8 & 82.95 & \textbf{+0.17} \\
\Xhline{1\arrayrulewidth} 
 Evol-Q & R & 4/8 & 67.29 & 4/8 & \textbf{77.06} & 4/8 & 82.63 & N/A \\
 \textbf{CLAMP-ViT (Ours)} & S & 4/8 & \textbf{69.93} & 4/8 & 77.03 & 4/8 & 82.69 & \textbf{+0.88} \\
 \Xhline{1\arrayrulewidth}
  \textbf{CLAMP-ViT (Ours)} & S & MP$_{4.9}$/MP$_{6.2}$ & {71.69} & MP$_{4.7}$/MP$_{5.9}$  & {79.43} & MP$_{4.8}$/MP$_{6.1}$ & \textbf{82.98} & - \\

 \Xhline{2\arrayrulewidth}
\end{tabular}
}
\vspace{-10pt}
\label{tab:compare_evol_q}
\end{table}
\endgroup

\subsection{Comparison with Evol-Q}
Since Evol-Q \cite{frumkin2023jumping}, does not quantize a model starting from a FP model and instead requires a pre-quantized model, we excluded including Evol-Q for comparison in the main paper since all baselines fully quantize a model from FP. We now show the comparison with Evol-Q and CLAMP-ViT for image classification task (Evol-Q is applicable only to image classification because of the nature of the global contrastive loss) in \cref{tab:compare_evol_q}. CLAMP-ViT fully quantizes the ViT from FP and calibrates on only 32 synthetic images. In contrast,  Evol-Q require 1000 calibration images from the original training and starts from an already quantized model. As shown in the \cref{tab:compare_evol_q}, despite significantly fewer fine-tuning samples that too at the absence of original images, CLAMP-ViT outperforms  Evol-Q averaged across different quantization scenarios with different ViT families.

\subsection{Additional Ablations}
\begin{wraptable}{r}{0.5\textwidth}
\vspace{-12mm}
		\caption{Ablation showing effects of our synthetic data on Evol-Q for W4/A8 quantization.}
		\label{table_evol_new}
		\centering
		\begin{tabular}{cccc}
\Xhline{2\arrayrulewidth}
Model & Data & \# Images & Top-1 \\
\Xhline{2\arrayrulewidth}
 \multirow{2}{*}{DeiT-T} & R & 1000 & {67.29} \\
 & S & 32 & \textbf{67.29} \\
 \Xhline{1\arrayrulewidth}

  \multirow{2}{*}{Swin-S} & R & 1000 & \textbf{82.63} \\
 & S & 32 & 82.51 \\
 
 \Xhline{2\arrayrulewidth}
\end{tabular}
\vspace{-7mm}
\end{wraptable}
To further study the effectiveness of our generated synthetic data for quantization and assess wider applicability of our data to other methods, we conduct an experiment wherein we replace Stage 2 in CLAMP-ViT with Evol-Q \cite{frumkin2023jumping}. This is straightforward as even Evol-Q uses a version of evolutionary search for adjusting scale factors. We report the W4/A8 quantization results in \cref{table_evol_new} where `R' signifies real-data of 1000 calibration images and corresponds to the standard Evol-Q and `S' signifies synthetic data of batch size 32 and corresponds to our modified version. We can infer from \cref{table_evol_new} that using our generated synthetic data we are able to closely match the original Evol-Q with 1000 real-world images. This experiment provides conclusive proof of the ability of the generalizability of our generated data and potential to match quantization performance on real world data.

We also demonstrate in \cref{fig:iteration} with the DeiT-S model that altering too many layer parameters simultaneously during quantization causes search instability and poor convergence. In contrast, a layer-wise search approach, as used in CLAMP-ViT's quantization framework, achieves optimal performance.

%
%

\bibliographystyle{splncs04}
\bibliography{main}


\end{document}